\documentclass{article}
\usepackage{mlspconf,amsmath,amssymb,graphicx}
\usepackage{amsfonts}
\usepackage{url}
\usepackage{algorithm}
\usepackage{algorithmicx}
\usepackage{algpseudocode}
\usepackage{booktabs}
\usepackage{threeparttable}
\usepackage{multirow}
\usepackage{caption,subcaption}
\usepackage{array}

\graphicspath{{figure/}}
\newcommand{\tabincell}[2]{
    \begin{tabular}{@{}#1@{}}#2\end{tabular}
}
\newcommand{\eqvspace}{
    \vspace{0mm}
}

\title{BALSON: BAYESIAN LEAST SQUARES OPTIMIZATION \\WITH NONNEGATIVE L1-NORM CONSTRAINT}

\name{Jiyang Xie$^1$, Zhanyu Ma$^{1,*}$, Guoqiang Zhang$^2$, Jing-Hao Xue$^3$, Jen-Tzung Chien$^4$, Zhiqing Lin$^1$, Jun Guo$^1$}
\address{\small$^1$Pattern Recognition and Intelligent Systems Lab., Beijing University of Posts and Telecommunications, China\\
\small$^2$School of Computing and Communications, University of Technology Sydney, Australia\\
\small$^3$Department of Statistical Science, University College London, United Kingdom\\
\small$^4$Department of Electrical and Computer Engineering, National Chiao Tung University, Taiwan\thanks{$^*$Corresponding author.}}

\begin{document}
%

\maketitle
\begin{abstract}
A Bayesian approach termed BAyesian Least Squares Optimization with Nonnegative $L_1$-norm constraint (BALSON) is proposed. The error distribution of data fitting is described by Gaussian likelihood. The parameter distribution is assumed to be a Dirichlet distribution. With the Bayes rule, searching for the optimal parameters is equivalent to finding the mode of the posterior distribution. In order to explicitly characterize the nonnegative $L_1$-norm constraint of the parameters, we further approximate the true posterior distribution by a Dirichlet distribution. We estimate the statistics of the approximating Dirichlet posterior distribution by sampling methods. Four sampling methods have been introduced. With the estimated posterior distributions, the original parameters can be effectively reconstructed in polynomial fitting problems, and the BALSON framework is found to perform better than conventional methods.
\end{abstract}
\vspace{-2mm}
\vspace{-7mm}
\section{Introduction}
\label{sec:intro}
\vspace{-4mm}

In machine learning and statistics, optimization methods, including Newton's method \cite{nocedal06}, quasi-Newton method \cite{nocedal06}, sequence quadratic programming (SQP) method \cite{gill05}, gradient descent method \cite{goodfellow16}, interior-point (IP) method \cite{byrd00}, and Bayesian methods \cite{chien15,park08,ohara2009a}, are widely applied. The least squares optimization (LSO), which is one of the unconstrained optimization problems, includes the residual sum of squares (RSS) errors as the objective function. This optimization can be proved and solved by proven algorithms with low computational complexity \cite{boyd04,schm05}. On this foundation, introduction of constraint conditions is beneficial to achieve numerical stability and increase predictive performance \cite{schm05}.

Sparsity is a common constraint to make the objective function depend on only a small number of model parameters. $L_0$- and $L_1$-norm regularizations are the commonly used constraints for sparsity. $L_0$-norm, denoted as $\left\|\cdot\right\|_0$, which can be defined as the number of non-zero elements in the parameter vector, performs the most precise sparsity of parameters, yet is difficult to implement in practice. $L_1$-norm, denoted as $\left\|\cdot\right\|_1$, which can be defined as the sum of the absolute values of the elements in a parameter vector, performs a strong sparsity constraint to the vector, and is convenient to be applied. With the constraint of $L_1$-norm regularization, the sparse representation \cite{cheng13}, the nonlinear programming \cite{sra11}, and nonlinear time series prediction \cite{koca14} are applied.

In addition to the aforementioned methods, solution under Bayesian framework is an alternative solution. With the probabilistic interpretation, the LSO problem (\emph{i.e.}, the RSS objective function) is usually treated as Gaussian likelihood, and the constraint is considered as prior distribution. Combining the likelihood function with the prior distribution and with the Bayes theorem, finding the optimal solution to the constrained LSO problem is then equivalent to calculating the mode of the posterior distribution. This is a maximum \emph{a posteriori} (MAP) solution to the constrained LSO problem. For example, with the $L_1$-norm constraint, the prior distribution is usually assumed to be a Laplacian \cite{park08,moha12,will95}. Chien \cite{chien15} proposed a Bayesian framework based on the Laplace prior of model parameters for sparse representation of sequential data. Finding the mode of the posterior distribution for Gaussian likelihood and Laplacian prior can solve the sparse optimization problem with numerical simulation.

There exists another type of regularization with nonnegative $L_1$-norm constraint, \emph{i.e.}, the regularization term contains nonnegative elements only \cite{schm05}. Nonnegative constraint plays an important role for solving the general nonnegative linear or nonlinear programming problems in physics (for example, fluid physics) \cite{naga11} and engineering applications (for example, hyperspectral image processing, audio processing, web documents analysis, and bioinformatics data processing) \cite{chien15,cheng13,satio15}. In this case, Laplacian assumption cannot describe the constraint well as it has negative support.

In this paper, \emph{we propose a Bayesian learning framework to solve this LSO problem with nonnegative $L_1$-norm constraint}. In order to capture the distribution for the constraint term precisely, the Dirichlet distribution is applied. Bhattacharya et al. \cite{bhat15} introduced Dirichlet-Laplace priors for optimal shrinkage. Sato et al. \cite{sato07} used the parametric mixture model with Dirichlet prior for knowledge discovery of multiple-topic document. Since combining the Gaussian likelihood for the model residual and the Dirichlet prior for the model parameters does not lead to a Dirichlet posterior, this paper approximates the posterior distribution with a Dirichlet distribution. Therefore, the optimal solution to LSO problem with nonnegative $L_1$-norm constraint can be solved by finding the mode of the approximating Dirichlet posterior distribution.

However, there is no analytically tractable solution to find the parameters of the aforementioned approximating Dirichlet posterior distribution. Sampling is an available method to analyze the statistical property of posterior distributions. Girolami et al. \cite{giro05} used the importance sampling to calculate the corresponding moments with respect to the posterior distribution over the Dirichlet parameters. In addition to the importance sampling, other sampling methods including the rejection sampling \cite{bishop06} can also be applied. We propose an approach, called BAyesian Least Squares Optimization for Nonnegative $L_1$-norm constrain (BALSON), which utilizes sampling methods to approximate the required statistical properties (including mode) of posterior distributions.

\vspace{-4mm}
\section{Bayesian Least Squares Optimization with Nonnegative L1-Norm Constraint}
\label{sec:method}
\vspace{-3mm}

\subsection{Problem Formulation}
\label{ssec:formulation}
\vspace{-2mm}

A LSO problem with nonnegative $L_1$-norm constraint can be defined as
\eqvspace
\begin{equation}
    \small
    \begin{array}{rl}
        &\min_{\boldsymbol{\theta}}\left\|y-f(\boldsymbol{x};\boldsymbol{\theta})\right\|_2^2,\\
        \text{s.t. }&\sum_i\theta_i\le C,\theta_i\ge0, i=1,\cdots,K
    \end{array}\label{eq:lso2}
    \eqvspace
\end{equation}
\noindent where $\boldsymbol{x}$ is the input of the model, $f(\cdot;\boldsymbol{\theta})$ is the model function with parameters $\boldsymbol{\theta}=\left[\theta_1,\cdots,\theta_K\right]^\text{T}$, $y$ is the observed target value, and $C$ is a constant. In this case, the optimization problem is equivalent to
\eqvspace
\begin{equation}
    \small
    \begin{array}{rl}
        &\min_{\boldsymbol{\theta}}\left\|y-f(\boldsymbol{x};\boldsymbol{\theta})\right\|_2^2+\lambda\boldsymbol{1}^{\text{T}}\boldsymbol{\theta},\\
        \text{s.t. }&\theta_i\ge0,i=1,\cdots,K
    \end{array}\label{eq:lso3}
    \eqvspace
\end{equation}
\noindent where $\lambda$ is a Lagrangian multiplier, and $\boldsymbol{1}$ is a column vector of ones. As we know, transformation $\ln t$ requires a nonnegative variable $t$ which is suitable for our nonnegative constraint. Thus, we can introduce a log-barrier penalty with a group of positive hyperparameters $\mu_i,i=1,\cdots,K$ to deal with the nonnegative constraint on the $\theta_i,i=1,\cdots,K$, and the condition $\theta_i\ge0$ can be replaced by $\sum_{i=1}^{K}\mu_i\ln\theta_i=M$, where $M$ is a constant \cite{schm05}.

\begin{figure}
   \centering
        \includegraphics[width=0.38\textwidth]{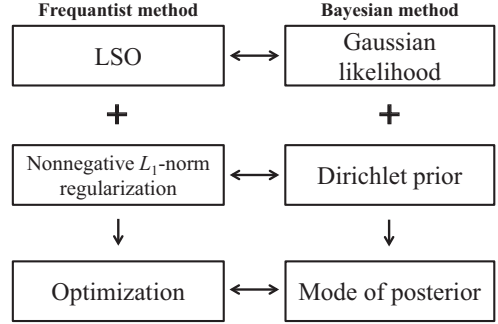}\\
    \vspace{-4mm}
   \caption{\small The relation between the LSO problem with nonnegative $L_1$-norm constraint and the proposed Bayesian framework.}\label{fig:relationship}
   \vspace{-6mm}
\end{figure}

With nonnegative $L_1$-norm constraint, the problem is presented as
\eqvspace
\begin{equation}
    \small
    \min_{\boldsymbol{\theta}}\underbrace{\left\|y-f(\boldsymbol{x};\boldsymbol{\theta})\right\|_2^2}_{\mathbb A}+\underbrace{\lambda\boldsymbol{1}^{\text{T}}\boldsymbol{\theta}-\sum_{i=1}^{K}\mu_i\ln\theta_i}_{\mathbb B},\label{eq:nneglasso}
    \eqvspace
\end{equation}
\noindent where $\{\mu_i\}_{i=1}^K$ are the Lagrangian multipliers. It is assumed that the model residual $e=y-f(\boldsymbol{x};\boldsymbol{\theta})$ follows the Gaussian distribution with zero mean and unit variance. Therefore, term $\mathbb A$ in \eqref{eq:nneglasso} can be considered as the negative logarithm of a Gaussian likelihood with zero mean and unit variance up to a constant difference as
\eqvspace
 \begin{equation}
    \small
    \mathbb{A}=-\ln{\mathcal{N}(y-f(\boldsymbol{x};C\boldsymbol{\omega});0,1)}+C_{\mathbb A},\label{eq:parta}
    \eqvspace
 \end{equation}
\noindent where $\boldsymbol{\omega}=\frac{\boldsymbol{\theta}}{C}$. Moreover, term $\mathbb B$ can be considered as the negative logarithm of a Dirichlet prior up to a constant difference as
\eqvspace
\begin{equation}
    \small
    \mathbb{B}=-\ln{\text{Dir}(\boldsymbol{\omega};\boldsymbol{\alpha})}+C_{\mathbb B}.\label{eq:partb}
    \eqvspace
\end{equation}
\noindent A Dirichlet distribution $\text{Dir}(\boldsymbol{\omega};\boldsymbol{\alpha})$ with parameter vector $\boldsymbol{\alpha}=[\alpha_1,\alpha_2,\cdots,\alpha_{K}]^{\text{T}},\alpha_i>0,i=1,\cdots,K$ is defined as
\eqvspace
\begin{equation}
    \small
    \text{Dir}(\boldsymbol{\omega};\boldsymbol{\alpha})=\frac{1}{B(\boldsymbol{\alpha})}\prod_{i=1}^{K}\omega_i^{\alpha_i-1},\label{eq:dirichlet}
    \eqvspace
\end{equation}
\noindent where order $K\ge2$, $\sum_{i=1}^{K}\omega_i=1$ and $\omega_i\ge0,i=1,\cdots,K$, and $B(\boldsymbol{\alpha})$ is the multivariate beta function as a normalization constant. It is worthy to note that this $L_1$-norm constraint is guaranteed by the definition of the Dirichlet, and $\boldsymbol{1}^{\text{T}}\boldsymbol{\theta}=\sum_i\theta_i=C\cdot\sum_i\omega_i=C$ is a constant. Hence, the term $\lambda\boldsymbol{1}^{\text{T}}\boldsymbol{\theta}$ in $\mathbb B$ can be neglected. Then, we can convert the original minimization problem in \eqref{eq:nneglasso} to a maximization problem as
\eqvspace
\begin{equation}
    \small
    \max_{\boldsymbol{\omega}}\left[\ln{\mathcal{N}(y-f(\boldsymbol{x};C\boldsymbol{\omega});0,1)}+\ln{\text{Dir}(\boldsymbol{\omega};\boldsymbol{\alpha})}\right].\label{eq:objfunc}
    \eqvspace
\end{equation}
\noindent The maximization operation in \eqref{eq:objfunc} is equivalent to calculating the mode of the posterior distribution characterized by a Gaussian likelihood described in \eqref{eq:parta} and a Dirichlet prior distribution in \eqref{eq:partb}. The relation between the LSO problem with nonnegative $L_1$-norm constraint and the proposed Bayesian framework is shown in Figure \ref{fig:relationship}.

\vspace{-4mm}
\subsection{Implementation Procedure}
\label{ssec:impleproced}
\vspace{-2mm}

The true posterior distribution characterized by a Gaussian likelihood and a Dirichlet prior distribution has a complex form which is not feasible in practice. In order to preserve the nonnegative $L_1$-norm constraint, we further \emph{assume} that the posterior distribution follows a Dirichlet distribution. Although under such conditions there is no analytically tractable solution to get the parameters, we can approximate the actual posterior distribution by matching the moments \cite{franke15} in the approximating Dirichlet posterior $\text{Dir}(\boldsymbol{\omega};\boldsymbol{\alpha^*})$. A numerical solution is proposed here to estimate the moments in the Dirichlet posterior by sampling methods. The first and second order moments of the Dirichlet posterior distribution are
\eqvspace
\begin{eqnarray}
    \small
    \begin{array}{rcl}
        \mathbb{E}[\omega_i]&=&\frac{\alpha_i^*}{\alpha_0^*},\\
        \text{Var}[\omega_i]&=&\mathbb{E}[(\omega_i-\mathbb{E}[\omega_i])^2]=\frac{\alpha_i^*(\alpha_0^*-\alpha_i^*)}{(\alpha_0^*)^2(\alpha_0^*+1)},
    \end{array}\label{eq:moments}
\end{eqnarray}
\noindent where $\mathbb{E}[\omega_i]$ and $\text{Var}[\omega_i]$ can also be estimated by mean value and variance of the $i^{th}$-dimensional samples respectively, $i=1,\cdots,K$, and $\alpha_0^*=\sum_i\alpha_i^*$. Then, $\boldsymbol{\alpha^*}$ can be computed directly by solving the linear equations in \eqref{eq:moments}. With the estimated parameters $\boldsymbol{\alpha^*}$, the optimal $\boldsymbol{\omega^*}$ to the LSO problem with nonnegative $L_1$-norm constraint can be obtained by computing the mode of the Dirichlet posterior distribution directly as
\eqvspace
\begin{equation}
    \small
    \omega^*_i=\left\{
    \begin{array}{cl}
    \frac{\alpha_i^*-1}{\sum_{j|\alpha_j^*>1}\alpha_j^*-K^*} & \alpha_i^*>1 \\
    0 & \text{otherwise}\label{eq:mode}
    \end{array}
    \right.,
    \eqvspace
\end{equation}
\noindent where $K^*$ is the number of $\alpha_j^*>1,j=1,\cdots,K$. From \eqref{eq:dirichlet}, we can observe that, when $\alpha_i^*$ is smaller than $1$, the marginal distribution on the $i^{th}$ dimension is with a convex shape. Hence, the mode of this dimension is at $0$ where the likelihood is infinity, and the other dimensions should be normalized as in \eqref{eq:mode}. This ensures the sparse property of nonnegative $L_1$-norm constraint. Then, the optimal $\boldsymbol{\theta}^*$ in the original problem \eqref{eq:lso2} can be computed by $\boldsymbol{\omega}^*$ as
\eqvspace
\begin{equation}
    \small
    \boldsymbol{\theta}^*=C\boldsymbol{\omega}^*.\label{eq:optimalw}
    \eqvspace
\end{equation}

The algorithm of BALSON is summarized in Algorithm \ref{alg:method}. Four sampling methods have been applied in the algorithm: rejection sampling (RS), importance sampling (IS), rejection sampling importance resampling (RSIRS) and importance sampling importance resampling (ISIRS), which are described in Section \ref{ssec:sampling}.

\vspace{-5mm}
\section{Bayesian Inference and Interpretation}
\label{sec:bayesinferandinterp}

\vspace{-4mm}
\subsection{Sampling Solutions}
\label{ssec:sampling}

\vspace{-2mm}
\subsubsection{Rejection Sampling}
\label{sssec:rs}
\vspace{-2mm}

The rejection sampling (RS) method allows us to generate enough acceptable samples from relatively complex target distributions $p(z)$, and reject samples which do not satisfy the target distribution, subject to certain constraints \cite{bishop06}. A simpler proposal distributions $q(z)$, such as a Gaussian or Dirichlet distribution, and a constant $k$ whose value is selected to satisfy $kq(z) \ge p(z)$ for all $z$ are needed to draw samples easier. As an example, generating a one-dimensional sample $z_0$ needs three steps: (1) Generate a number $z_0$ from the distribution $q(z)$; (2) Generate a number $u_0$ from the uniform distribution over $[0, kq(z_0)]$; (3) Accept $z_0$ if $u_0 \le p(z_0)$, otherwise reject $z_0$. After having drawn $L$ (enough) samples, the moments can be calculated.

\begin{algorithm}[!t]
        \caption{BALSON}
        \label{alg:method}
        \begin{algorithmic}[1]
            \Require $\boldsymbol{x}$: input data; $\boldsymbol{y}$: observed target data; $K$: dimension of model parameters
            \Ensure $\boldsymbol{\theta}^*$: estimated model parameters
            \State Initial values: $\boldsymbol{\alpha}$: parameter vector of Dirichlet prior; $L$: number of sampling points.
            \State Sample from objective function in \eqref{eq:objfunc} using RS, IS, RSIRS or ISIRS and obtain $L$ samples (and their importance weights in IS and IRS).
            \State Compute the estimated parameters of Dirichlet posterior $\boldsymbol{\alpha}^*$ with the moments of $L$ samples in \eqref{eq:moments}.
            \State Compute the mode $\boldsymbol{\omega}^*$ with the method in \eqref{eq:mode}.
            \State Compute the optimal $\boldsymbol{\theta}^*$ with the method in \eqref{eq:optimalw}.
        \end{algorithmic}
\end{algorithm}

\vspace{-4mm}
\subsubsection{Importance Sampling}
\label{sssec:is}
\vspace{-2mm}

The importance sampling (IS) method allows us to sample from $p(z)$ only to approximate the moments instead of drawing samples \cite{bishop06}. All samples drawn from $q(z)$ directly are accepted and weighted. The normalized importance weights $\boldsymbol{\tilde{r}}=\{\tilde{r}_1,\cdots,\tilde{r}_L\}$ and the corresponding samples $\boldsymbol{z}=\{z_1,\cdots,z_L\}$ are used to calculate the moments.

\vspace{-4mm}
\subsubsection{Rejection Sampling Importance Resampling}
\label{sssec:rsirs}
\vspace{-2mm}

The rejection sampling importance resampling (RSIRS) is a combination of RS and IS methods. Firstly, we run RS one time and estimate parameter $\boldsymbol{\alpha^{(0)}}$ of the approximating Dirichlet posterior distribution. Then, the IS is implemented $R$ rounds and the parameters $\{\boldsymbol{\alpha^{(1)}},\cdots,\boldsymbol{\alpha^{(R)}}\}$ are estimated. In the $i^{th}$ iteration of IS, the estimated parameter $\boldsymbol{\alpha^{(i)}}$ of Dirichlet posterior is approximated by combining Gaussian likelihood of data and Dirichlet prior with the parameter $\boldsymbol{\alpha^{(i-1)}}$.

\vspace{-4mm}
\subsubsection{Importance Sampling Importance Resampling}
\label{sssec:isirs}
\vspace{-2mm}

Similar to RSIRS, the importance sampling importance resampling (ISIRS) is a combination of IS methods which will work several iterations. The only difference with RSIRS is that the first step is IS method instead of RS method.

\vspace{-4mm}
\subsection{Relationship with Bayesian LASSO}
\label{ssec:relation}
\vspace{-2mm}

Bayesian LASSO method is a popular Bayesian framework combining Gaussian likelihood and Laplace prior for sparse representation of model parameters \cite{park08}. Moreover, MAP problem in the Bayesian LASSO method can be converted to a negative logarithm form as a LSO problem with $L_1$-norm constraint as
\eqvspace
\begin{equation}
    \small
    \begin{array}{rl}
        &\min_{\boldsymbol{\tilde{\theta}}}\left\|y-f(\boldsymbol{x};\boldsymbol{\tilde{\theta}})\right\|_2^2,\\
        \text{s.t. }&\sum_i\left|\tilde{\theta}_i\right|\le C
    \end{array}\label{eq:lso}
    \eqvspace
\end{equation}
\noindent where $\boldsymbol{\tilde{\theta}}=\left[\tilde{\theta}_1,\cdots,\tilde{\theta}_{\tilde{K}}\right]^\text{T}$ are model parameters.

Following the approach in \cite{tibshirani96}, we extend the real vector $\boldsymbol{\tilde{\theta}}$ into a nonnegative real vector of $K$ ($K=2\tilde{K}$) dimensions $\boldsymbol{\theta}\triangleq\left[(\boldsymbol{\tilde{\theta}}^+)^{\text{T}},(\boldsymbol{\tilde{\theta}}^-)^{\text{T}}\right]^{\text{T}}$ as
\eqvspace
\begin{align}
    \small
    \theta_i=\tilde{\theta}_i^+&=
    \begin{cases}
        \tilde{\theta}_i&\tilde{\theta}_i>0\\
        0&\text{otherwise}
    \end{cases},\\
    \theta_{i+\tilde{K}}=\tilde{\theta}_i^-&=
    \begin{cases}
        -\tilde{\theta}_i&\tilde{\theta}_i<0\\
        0&\text{otherwise}
    \end{cases},\label{eq:wi}
    \eqvspace
\end{align}
\noindent where $\tilde{\theta}_i^+$ and $\tilde{\theta}_i^-$ are the elements of $\boldsymbol{\tilde{\theta}}^+$ and $\boldsymbol{\tilde{\theta}}^-$, respectively, and $\tilde{\theta}_i^+\ge0$, $\tilde{\theta}_i^-\ge0$, $i=1,\cdots,\tilde{K}$, which means that $\boldsymbol{\theta}$ is a nonnegative vector. Therefore,  $\tilde{\theta}_i$ has the relationship with $\tilde{\theta}_i^+$ and $\tilde{\theta}_i^-$ as
\begin{equation}
    \small
    \tilde{\theta}_i\triangleq\tilde{\theta}_i^+-\tilde{\theta}_i^-.\label{eq:widecomposed}
\end{equation}
\noindent It is noticed that $|\tilde{\theta}_i|=\tilde{\theta}_i^++\tilde{\theta}_i^-$.Thus, the $L_1$-norm constraint $\sum_i|\tilde{\theta}_i|\le C$ can be represented as $\sum_i(\tilde{\theta}_i^++\tilde{\theta}_i^-)=\sum_i\theta_i\le C$, which is a nonnegative $L_1$-norm constraint.

According to the aforementioned approach, the Bayesian LASSO method can also be solved by our BALSON framework. Moreover, we can apply BALSON framework to the LSO problem with both nonnegative and common $L_1$-norm constraints, but the Bayesian LASSO method can only solve the LSO problem with common $L_1$-norm constraint, which means that the proposed BALSON framework is more general than the Bayesian LASSO method.

\begin{figure}[!t]
  \centering
  \includegraphics[width=0.48\textwidth]{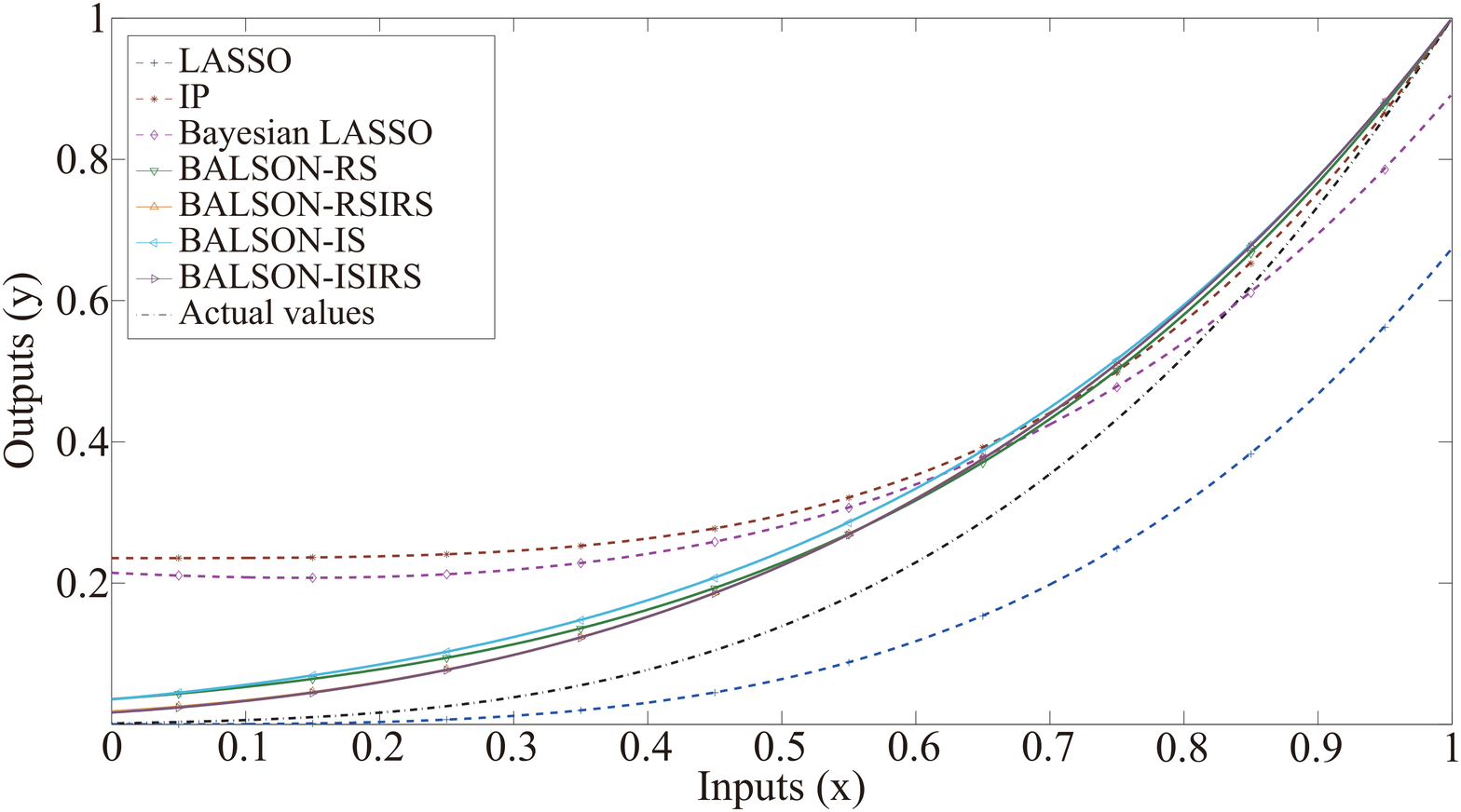}
  \vspace{-8mm}
  \caption{\small Actual and fitting curves using different methods.}\label{fig:polyfitcurve}
  \vspace{-6mm}
\end{figure}

\vspace{-5mm}
\section{Experimental Result and Discussion}
\label{sec:experiment}
\vspace{-3mm}

\subsection{Polynomial Fitting Problem}
\label{ssec:results}
\vspace{-2mm}
Bayesian polynomial curve fitting is an important problem in signal processing for its excellent performance on standard denoising and speech segmentation problems \cite{punskaya02,fearn05}. We apply the proposed BALSON framework to solve the polynomial fitting problem for illustrative purposes. The polynomial fitting problem aims to fit the $K$-dimensional polynomial parameters $\boldsymbol{\theta}=[\theta_1,\cdots,\theta_K]^\text{T}$, which can be expressed as
\eqvspace
\begin{equation}
    \small
    f(x;\boldsymbol{\theta})=\boldsymbol{\theta}^\text{T}\Phi(x),\label{eq:polyfit}
    \eqvspace
\end{equation}
\noindent where $x$ is the input scalar variable and the output is a scalar as well, and $\Phi(x)=\left[1,x,\cdots,x^{K-1}\right]^\text{T}$ is a polynomial kernel as the input vector variable in \eqref{eq:objfunc}. It is worth to note that, although we take scalar input and output as an example, the proposed method can also be applied to vector input and vector output.

Three methods have been implemented as reference methods. These methods can be categorized into two classes. One class of method is the frequentist method, which contains the least absolute shrinkage and selection operation (LASSO) \cite{tibshirani11} and the interior-point (IP) \cite{byrd00} algorithm. Another class of method belongs to Bayesian methods, from which we select the Bayesian LASSO \cite{park08}. The proposed BALSON (with four different implementations named as BALSON-RS,  BALSON-IS,  BALSON-RSIRS, and  BALSON-ISIRS, respectively) has been compared with the aforementioned methods, and the performance has been evaluated in terms of mean squared error (MSE) and sparsity. Here, MSE is defined as
\begin{equation}
    \small
    \text{MSE}=\frac{1}{N_{te}}(\boldsymbol{y}-\boldsymbol{\hat{y}})^{\text{T}}(\boldsymbol{y}-\boldsymbol{\hat{y}}),\label{eq:msefit}
    \eqvspace
\end{equation}
\noindent where $\boldsymbol{y}$ is the actual value vector, $\boldsymbol{\hat{y}}$ is the polynomial fitting value vector, and $N_{te}$ is the number of test points. Moreover, the sparsity \cite{hurley09} denotes the degree of sparseness of the estimated polynomial parameters $\boldsymbol{\hat{\theta}}$, which is defined as
\eqvspace
\begin{equation}
    \small
    \text{Sparsity}=\frac{\sqrt{K}-\frac{\left\|\boldsymbol{\hat{\theta}}\right\|_1}{\left\|\boldsymbol{\hat{\theta}}\right\|_2}}{\sqrt{K}-1},\label{eq:sparsity}
    \eqvspace
\end{equation}
\noindent where $\left\|\cdot\right\|_1$ and $\left\|\cdot\right\|_2$ denote $L_1$- and $L_2$-norm, respectively.

\begin{table}[!t]
    \vspace{-6mm}
    \centering
    \caption{\small MSE and sparsity using different methods.}\label{tab:polyfitmse}
    \vspace{-3mm}
    \footnotesize
    \small
    \setlength{\tabcolsep}{3pt}
    \begin{tabular}{ccccc}
    \toprule
    \multirow{2}[2]{*}{Method} & \multicolumn{2}{c}{Frequentist method} & \multicolumn{2}{c}{Bayesian method} \\
    \cmidrule{2-5}   & LASSO & IP    & \multicolumn{2}{c}{Bayesian LASSO} \\
    \midrule
    MSE& 0.0209  & 0.0257  & \multicolumn{2}{c}{0.0370 } \\
    Sparsity & 0.6660  & 0.5353  & \multicolumn{2}{c}{0.3833 } \\
    \midrule
    \multirow{2}[2]{*}{Method} & \multicolumn{4}{c}{BALSON (Bayesian method)} \\
    \cmidrule{2-5} & RS & RSIRS & IS & ISIRS \\
    \midrule
    MSE& 0.0143  & \textbf{0.0116 } & 0.0159  & 0.0158  \\
    Sparsity & 0.6751  & 0.8124  & 0.6785  & \textbf{0.8675 } \\
    \bottomrule
    \end{tabular}
    \vspace{-5mm}
\end{table}

Data of the polynomial fitting problem are generated for both training and test. In training step, a group of training points is generated by a $5$-dimensional polynomial parameter vector $\boldsymbol{\theta}=[0.0013,0.0380,0.0102,0.9082,0.0423] ^{\text{T}}$ (sparsity$=0.9200$) and $x$ obtained from $0$ to $1$ with fixed intervals according to the polynomial curve fitting model in \eqref{eq:polyfit}, and added noise which follows the Gaussian distribution with zero mean and unit variance. Then, in the test step, we pick test points on the estimated and actual curves at the same input values which are obtained from $0$ to $1$ with fixed intervals. These points are used to measure the differences between estimated and actual curves by MSE defined in \eqref{eq:msefit}. We set the number of training points $N_{tr}=100$ and number of test points $N_{te}=1000$, and $C=1$. All the seven methods (\emph{i.e.}, LASSO, IP, Bayesian LASSO, BALSON-RS, BALSON-IS, BALSON-RSIRS, BALSON-ISIRS) have been conducted $100$ times with the same data to obtain the distribution of MSE and sparsity. Figure \ref{fig:polyfitcurve} shows the comparisons among the estimated curves of different methods and the actual curve. The solid and dashed curves indicate estimated values of the proposed methods and the referred methods, respectively, and the dashdot curve indicates the actual values. It can be observed that the proposed methods yield the curves that are closer to the actual one than the referred methods.

\begin{figure}[!t]
    \vspace{-5mm}
    \centering
    \begin{subfigure}[t]{0.38\textwidth}
        \centering
        \includegraphics[width=1\textwidth]{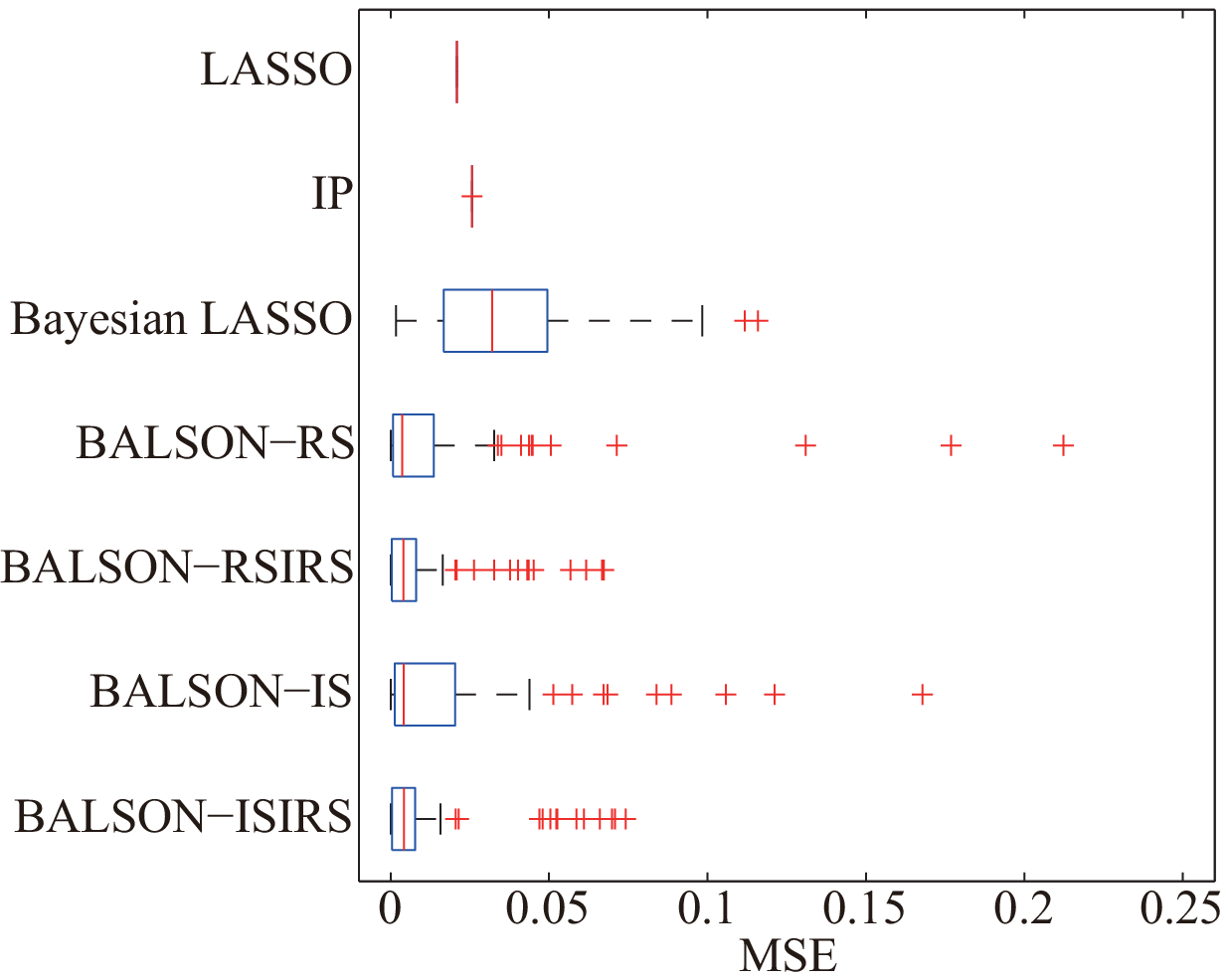}
        \vspace{-6.5mm}
        \subcaption{Distributions of MSE}
    \end{subfigure}
    \begin{subfigure}[t]{0.38\textwidth}
        \centering
        \includegraphics[width=1\textwidth]{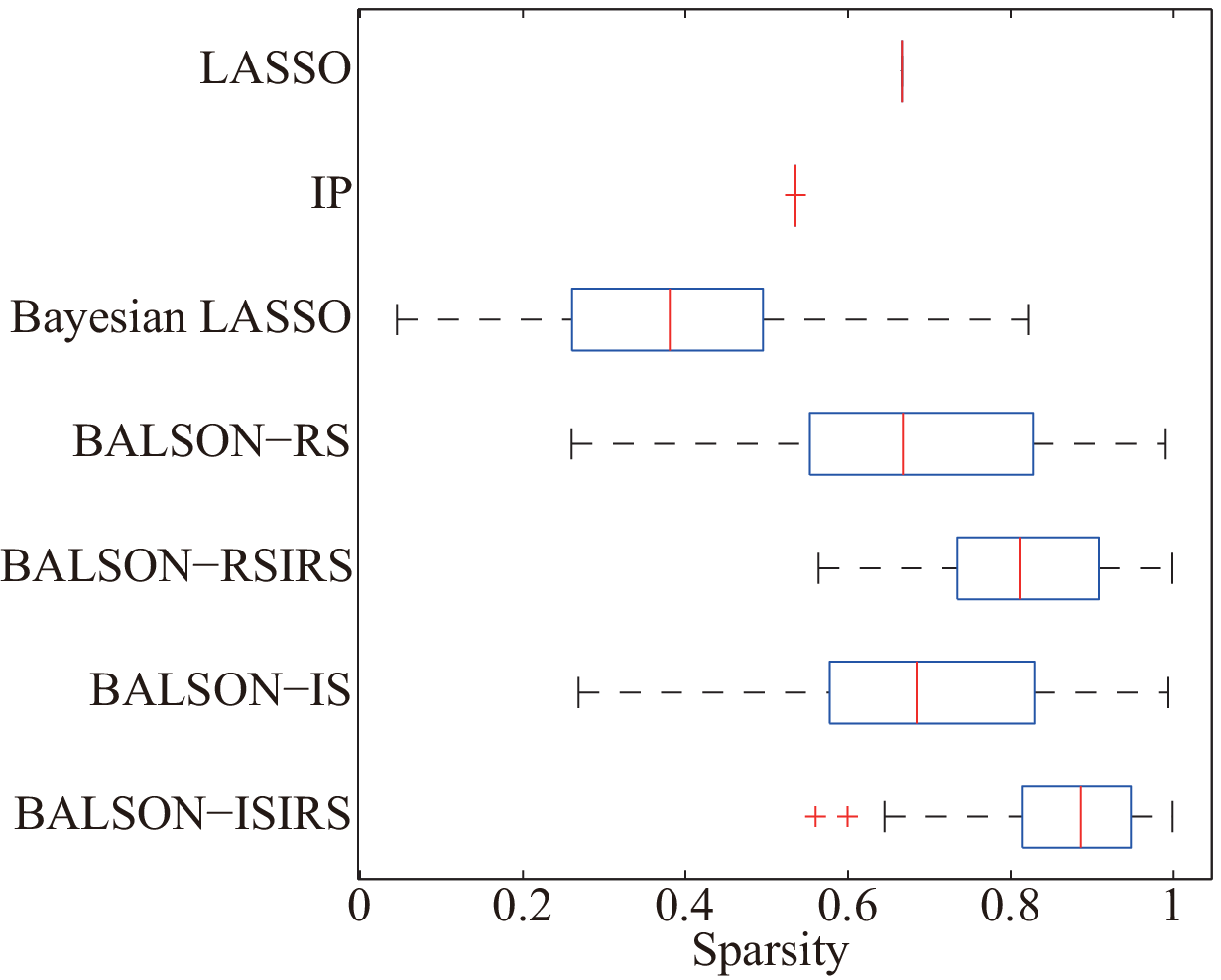}
        \vspace{-6.5mm}
        \subcaption{Distributions of sparsity}
    \end{subfigure}
    \vspace{-3.5mm}
    \caption{\small Boxplots for distributions of MSE and sparsity using different methods.}
    \label{fig:boxplot}
    \vspace{-7mm}
\end{figure}

To quantitatively evaluate the performance, the mean values of MSE and sparsity are shown in Table \ref{tab:polyfitmse}. The smallest MSE and the highest sparsity in Table \ref{tab:polyfitmse} are highlighted in bold. Moreover, the distributions of the MSE and the sparsity are shown in Figure \ref{fig:boxplot}(a) and \ref{fig:boxplot}(b), respectively. The proposed methods have smaller mean and median values of MSE than the referred methods, and the BALSON-RSIRS has the smallest mean value of MSE. Meanwhile, both of the BALSON-RSIRS and BALSON-ISIRS methods have higher mean and median values of sparsity which are larger than $0.8$. Generally speaking, the mean and median values of sparsity of the proposed methods are higher than the referred methods.

Moreover, paired \emph{t}-test on MSE and sparsity are conducted by setting the significance level as $0.05$, respectively. The corresponding \emph{p-values} are shown in Table \ref{tab:pmse} and \ref{tab:pspar}. Most of the \emph{p-values} of MSE computed between the proposed and the referred methods are smaller than $0.05$, except for the \emph{p-values} computed between BALSON-IS and LASSO, BALSON-ISIRS and LASSO, and BALSON-ISIRS and IP. In addition, most of the \emph{p-values} of sparsity are smaller than $0.05$, except for the \emph{p-values} computed between BALSON-RS and LASSO, IS and LASSO. Therefore, only BALSON-RSIRS method has statistically significant performance improvements from all the referred methods on both MSE and sparsity. It is worth to note that we have conducted several experiments with different parameter settings (\emph{i.e.}, different $\boldsymbol{\theta}$) and similar performances can be observed. Due to the limitation of space, we report only one example here.

\vspace{-5mm}
\subsection{Discussion}
\label{ssec:discussions}
\vspace{-3mm}

According to the results of the polynomial fitting experiments, the proposed methods have smaller MSE and higher sparsity. To specify, BALSON-RSIRS has the lowest MSE, and both BALSON-RSIRS and BALSON-ISIRS have higher sparsity than other methods. However, only BALSON-RSIRS method has statistically significant performance improvement from all the referred methods on both MSE and sparsity according to the results of the \emph{t}-test. Therefore, by considering both MSE and sparsity together, we suggest to apply BALSON-RSIRS method to perform simulation for the proposed Bayesian framework.

The computational complexity of BALSON-RS and BALSON-IS are only related to number of samples $L$, number of training data $N_{tr}$, and dimension of polynomial parameters $K$ when the rate of rejection is acceptable, which can be shown as $O(LK(K+N_{tr}))$. Thus, the computational complexity of BALSON-RSIRS and BALSON-ISIRS are $O(LK(K+N_{tr})R)$, where $R$ is the number of rounds in importance resampling.

\begin{table}[!t]
    \centering
    \vspace{-5mm}
    \caption{\emph{P-values} of MSE.}
    \vspace{-4mm}
    \footnotesize
    \small
    \begin{tabular}{lccc}
    \hline
          & LASSO & IP    & \tabincell{c}{Bayesian\\LASSO} \\
    \hline
    BALSON-RS    & 4.00E-02 & 4.86E-04 & 1.92E-06 \\
    BALSON-RSIRS & 2.43E-03 & 7.27E-06 & 1.52E-08 \\
    BALSON-IS    & 7.55E-02 & 6.13E-04 & 4.34E-07 \\
    BALSON-ISIRS & 3.74E-01 & 8.47E-02 & 3.93E-04 \\
    \hline
    \end{tabular}
  \label{tab:pmse}
\end{table}

\begin{table}[!t]
    \centering
    \vspace{-3.5mm}
    \caption{\emph{P-values} of sparsity.}
    \vspace{-4mm}
    \footnotesize
    \small
    \begin{tabular}{lccc}
    \hline
          &LASSO & IP    & \tabincell{c}{Bayesian\\LASSO} \\
    \hline
    BALSON-RS    & 6.18E-01 & 4.44E-14 & 1.85E-19 \\
    BALSON-RSIRS & 4.12E-24 & 4.33E-48 & 4.35E-38 \\
    BALSON-IS    & 5.15E-01 & 1.66E-13 & 4.75E-18 \\
    BALSON-ISIRS & 5.73E-37 & 3.83E-58 & 1.61E-44 \\
    \hline
    \end{tabular}
  \label{tab:pspar}
  \vspace{-7.5mm}
\end{table}

\vspace{-5mm}
\section{Conclusions}
\label{sec:conclusions}
\vspace{-4mm}

To solve the least squares optimization problem with nonnegative $L_1$-norm constraint, a novel Bayesian optimization method, BALSON, is proposed. The error distribution of data fitting is described by Gaussian likelihood while the Dirichlet prior and the approximating Dirichlet posterior were applied to satisfy the conjugate match requirement. As no analytically tractable solution exists, we estimate the properties of the Dirichlet posterior of the parameters by implementing sampling methods. With the estimated posterior distributions, the original parameters can be effectively reconstructed. In order to evaluate the performance of the proposed methods, the BALSON framework has been applied in the polynomial fitting problems. Compared with several referred methods, it achieved the best performance. In addition to the polynomial fitting problems, the proposed methods can be extended to other parameter estimation problems in many applications, such as hyperspectral image processing, audio signal processing, web documents analysis, and bioinformatics data processing. Future work will take into account optimizing by using variational Bayes methods \cite{ma15,ma2011bayesian,ma2014bayesian} for approximating the posterior distribution under the Kullback-Leibler (KL) divergence constraint. 
In addition, the relation between BALSON and compressive sensing will be explored.


\bibliographystyle{IEEEbib}

\end{document}